\documentclass[10pt,twocolumn,letterpaper]{article}

\usepackage{iccv}
\usepackage{times}
\usepackage{epsfig}
\usepackage{graphicx}
\usepackage{amsmath}
\usepackage{amssymb}

\usepackage{bbm}
\usepackage{booktabs}
\usepackage{epstopdf}
\usepackage{multirow}   
\usepackage{subfigure}  
\usepackage{algorithm}
\usepackage{algorithmic}
\usepackage{makecell}
\usepackage[misc]{ifsym}
\usepackage[accsupp]{axessibility} 
\usepackage[breaklinks=true,bookmarks=false]{hyperref}

\iccvfinalcopy 

\ificcvfinal\fi

\begin{document}

\title{Data-free Knowledge Distillation for Fine-grained Visual Categorization}

\renewcommand{\thefootnote}{\fnsymbol{footnote}}
\author{Renrong Shao$^1$,\quad Wei Zhang$^{1}$\footnotemark[1],\quad Jianhua Yin$^2$,\quad Jun Wang$^{1}$\footnotemark[1]\\
$^1$School of Computer Science and Technology, East China Normal University \\
$^2$School of Computer Science and Technology, Shandong University\\
{\tt\small \{roryshaw6613,zhangwei.thu,jhyinmail, wongjun\}@gmail.com}
}

\maketitle

\footnotetext[1]{Corresponding authors. This work was supported in part by National Natural Science Foundation of China under Grant (No. 62072182, No. 92270119, and No. 62172261) and Key Laboratory of Advanced Theory and Application in Statistics and Data Science, Ministry of Education.}
\renewcommand{\thefootnote}{\arabic{footnote}}

\ificcvfinal\thispagestyle{empty}\fi

\begin{abstract}
Data-free knowledge distillation (DFKD) is a promising approach for addressing issues related to model compression, security privacy, and transmission restrictions. 
Although the existing methods exploiting DFKD have achieved inspiring achievements in coarse-grained classification, 
in practical applications involving fine-grained classification tasks that require more detailed distinctions between similar categories, sub-optimal results are obtained. 
To address this issue, we propose an approach called DFKD-FGVC that extends DFKD to fine-grained visual categorization~(FGVC) tasks. 
Our approach utilizes an adversarial distillation framework with attention generator, mixed high-order attention distillation, and semantic feature contrast learning. Specifically, we introduce a spatial-wise attention mechanism to the generator to synthesize fine-grained images with more details of discriminative parts. 
We also utilize the mixed high-order attention mechanism to capture complex interactions among parts and the subtle differences among discriminative features of the fine-grained categories, paying attention to both local features and semantic context relationships. 
Moreover, we leverage the teacher and student models of the distillation framework to contrast high-level semantic feature maps in the hyperspace, comparing variances of different categories. We evaluate our approach on three widely-used FGVC benchmarks (Aircraft, Cars196, and CUB200) and demonstrate its superior performance. 
Code is available at \href{https://github.com/RoryShao/DFKD-FGVC.git}{https://github.com/RoryShao/DFKD-FGVC.git}
\end{abstract}

\section{Introduction}
Fine-grained visual categorization (FGVC) aims at distinguishing subcategories from father categories, e.g., subcategories of birds~\cite{welinder2010caltech}, aircraft~\cite{maji2013fine}, and cars~\cite{krause20133d}.
It has long been considered a more challenging issue than traditional image classification due to the subtle inter-class and large intra-class variations~\cite{wei2021fine}.
To distinguish subtle diversities, the current approaches commonly exploit deeper neural networks with elaborate designs~\cite{zhao2021graph, liu2020filtration, zhuang2020learning}  to excavate the discriminative features effectively.
Inevitably, the network becomes more and more complex, which leads to another problem, i.e., complicated networks are not easily deployed on embedded or mobile devices.
Besides, the training data of released pre-trained models are often unavailable due to transmission, privacy, or legal issues. 
For example, pre-trained models commonly need a large amount of data such as ImageNet~\cite{krizhevsky2012imagenet}. If the data is transmitted directly, a large amount of bandwidth is consumed. Moreover, some sensitive data such as e-commerce items or medical data are usually not directly accessible to the public due to intellectual property rights or privacy protection considerations. 
To obtain a lightweight model, recent research has made significant progress, including pruning~\cite{li2016pruning}, quantization~\cite{zhao2019improving,liu2021zero}, and knowledge distillation~\cite{hinton2015distilling}.
Among them, knowledge distillation (KD) is a popular and effective paradigm for model compression and knowledge transfer~\cite{hinton2015distilling}. It works by transferring knowledge from a cumbersome teacher network to a lightweight student network. Thanks to this separable architecture, it can also be used to solve privacy protection in data-free scenarios, which is called data-free knowledge distillation~(DFKD)~\cite{chen2019data} or zero-shot knowledge distillation~(ZSKD)~\cite{nayak2019zero}.

Fortunately, a series of DFKD methods have been proposed~\cite{chen2019data,nayak2019zero,micaelli2019zero,fang2019data,yin2020dreaming,fang2021contrastive}.
The existing approaches can be divided into two paradigms.
The first paradigm is based on the category distribution, which exploits the out distribution of teacher and student to optimize the student and generator, e.g., DFAL~\cite{chen2019data},  ZSKT~\cite{micaelli2019zero}, DFAD~\cite{fang2019data}, ZSKD~\cite{nayak2019zero}. Such a paradigm commonly fails to generate realistic samples due to the lack of semantic-related information, especially when it comes to complex samples.
The second paradigm is based on prior distribution, which exploits the prior information~(i.e., BatchNorm) to optimize synthetic images for distillation, e.g., MAD~\cite{domomentum},  CMI~\cite{fang2021contrastive}, DFQ~\cite{choi2020data}, ADI~\cite{yin2020dreaming}. This paradigm can produce realistic features and, therefore, gives the student a noticeable improvement.

Although the existing methods have achieved inspiring achievements in coarse-grained classification,  in practical applications, sub-optimal results are achieved due to the subtle variations widely found in different scenarios.
The main reasons for this situation are as follows:
Firstly, for FGVC tasks, the variances of the same category are more prominent than that of coarse-grained classification due to different factors, such as viewing angles, lighting, backgrounds, occlusion, etc.
Secondly, compared to coarse-grained categories, the feature discrepancies of different categories in FGVC are not obvious.
Besides, in the data-free scenario, the model can not access the raw data directly. For synthesized images, it is difficult for the teacher model to capture the subtle variances of discriminative features.
To our best knowledge, there are still no specialized data-free studies on fine-grained DFKD.
Therefore, this inspires us to explore this issue and tackle this task in a data-free scenario.

\begin{figure*}[!t]
  \centering
  \includegraphics[width=\linewidth]{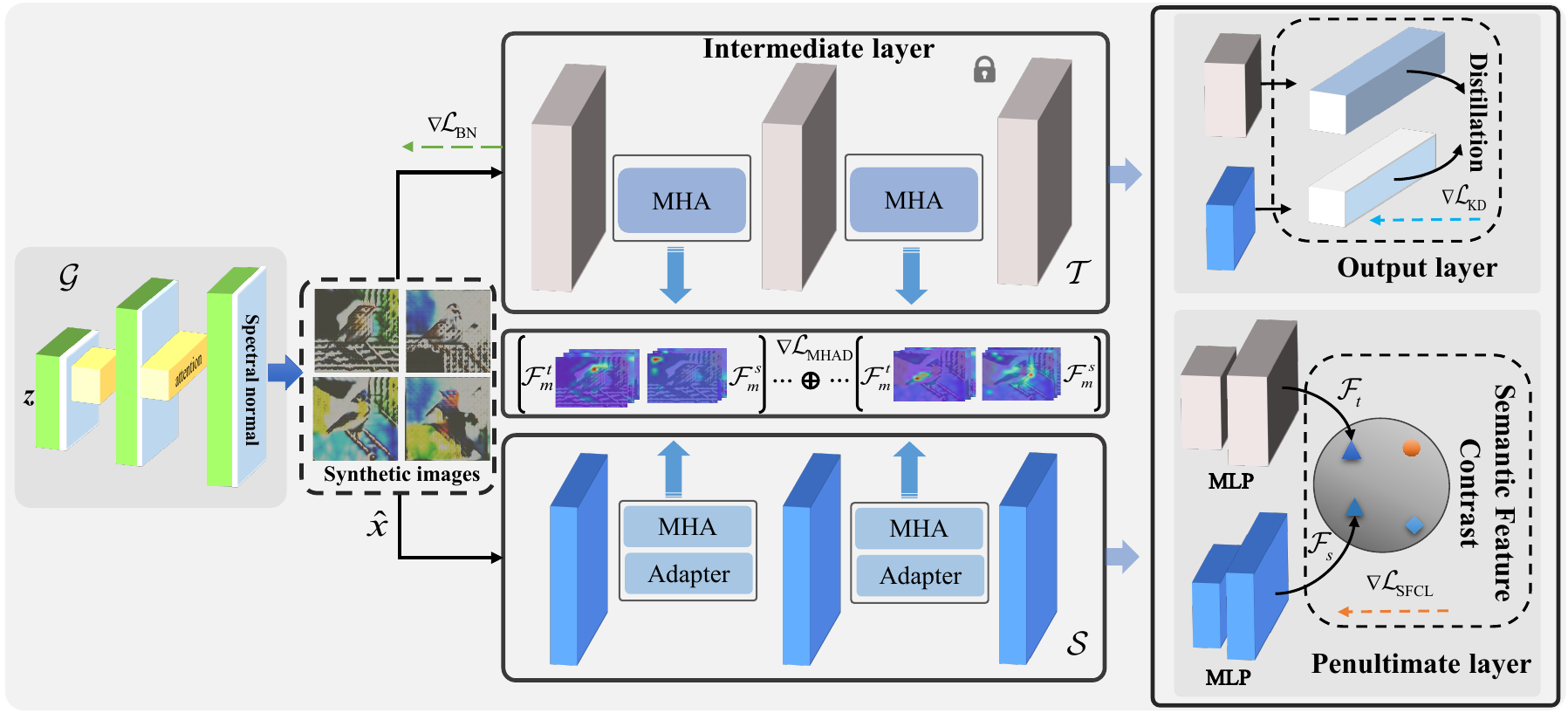}
  \vspace{-2mm}
  \caption{The whole framework of our approach. The \textbf{left}: The spatial attention module is plugged into each block of generator $\mathcal{G}$, which aims to focus on fine-grained semantic information from the whole process of noise $z$ to images $\hat{x}$.  The \textbf{intermediate}: At each block of teacher and student, the feature maps are extracted by the mixed high-order attention module to achieve MHAD.
  The \textbf{right}: In the penultimate layer, exploiting the MLP to map the high-level semantic features of teacher and student to a common hyperspace and compare the variances by SFCL.}
  \label{fig:framework}
  \vspace{-3mm}
\end{figure*}

In this paper, we tackle this issue by extending DFKD to fine-grained visual classification~(FGVC) tasks and propose an approach named DFKD-FGVC, which is achieved by exploiting the adversarial distillation framework with attention generator, mixed high-order attention distillation~(MHAD) and semantic feature contrast learning~(SFCL).
Concretely, as shown in Fig.~\ref{fig:framework}, to promote the generator to synthesize more fine-grained images, we exploit the generator with spatial-wise attention, which can help the generator synthesize the images with more details of discriminative parts.
Then, to fully mine the knowledge of discriminative features for student, we exploit the mixed high-order attention mechanism to capture complex interactions among parts and the subtle differences among discriminative features of the fine-grained categories, paying attention to both local features and semantic context relationships. 
Besides, to compare variances of different categories, we skillfully exploit the teacher and student model of distillation framework to contrast semantic feature maps in the hyperspace. 
To verify our approach, massive experiments are conducted on three fine-grained benchmarks, such as Aircraft, Cars196, and CUB200 to evaluate the effectiveness of our approach.

In a nutshell, our contributions are four-fold:
~1) We are the first to propose an approach for FGVC in the data-free distillation scenario, which aims to optimize the entire generation and distillation stages to focus on discriminative features. ~2)  To synthesize more fine-grained images for adversarial distillation, we employ the generator with spatial-wise attention, which motivates the generator to synthesize the images with more details of discriminative features. ~3)  Particularly, to effectively mine the potential semantic features and contextual relationships of the fine-grained categories, we provide two strategies, namely, MHAD and SFCL, both of which can promote the performance of DFKD from different dimensions. ~4) Extensive experiments demonstrate the effectiveness of our approach in the data-free setting, which achieves state-of-the-art performance on the mainstream FGVC benchmark datasets.

\section{Related Works}
\subsection{Fine-Grained Visual Categorization}
Fine-grained visual classification (FGVC)~\cite{welinder2010caltech,maji2013fine,krause20133d} is much more challenging than traditional classification tasks due to the inherently subtle intra-class object variations~\cite{wei2021fine,huang2016part}.
Benefiting from the recent development of neural networks, recent studies have moved from strongly supervised information with extra annotations such as bounding boxes~\cite{berg2013poof, zhang2014part, huang2016part} to weakly-supervised conditions with only category labels~\cite{zheng2017learning, ge2019weakly, wang2020graph}.
Current methods on FGVC can be roughly divided into localization-based methods ~\cite{ge2019weakly,wang2020graph} and attention-based methods~\cite{cai2017higher, ji2020attention, min2020multi}. The core for solving FGVC is to learn the discriminative features of objects in images.
However,  current approaches tackle this problem in the data-driven setting, few approaches consider this problem in the data-free setting.
Therefore, different from the above studies, we explore the FGVC tasks in the novel aspect of the data-free distillation scenario. 

\subsection{Attention Mechanism}
The attention mechanism stems from human vision, which exploits a sequence of partial glimpses and selectively focuses on salient parts to capture visual structure better.
In the field of computer vision, attention mechanism~\cite{wang2017residual,hu2018squeeze,woo2018cbam,parkbam} are mainly exploited to capture essential information in various tasks such as pedestrian re-identification~\cite{xu2018attention,ji2020attentiondriven,wu2021attention}, FGVC ~\cite{cai2017higher, min2020multi}, etc. For example,  ~\cite{wang2017residual} proposes the residual attention network for large-scale classification tasks. 
Then Hu et al.~\cite{hu2018squeeze} exploit a squeeze-and-excitation~(SE) block to compute channel-wise attention. CBAM~\cite{woo2018cbam} infers attention maps along two separate dimensions, i.e., channel and spatial. Similar to ~\cite{woo2018cbam}, BAM~\cite{parkbam} also exploits the 3D attention map inference into channel and spatial. 
In terms of tasks, spatial attention is well-suited to dense prediction tasks such as semantic segmentation and object detection~\cite{guo2022attention}, while channel attention is a good choice for image classification. However, only exploiting spatial attention or channel attention is coarse, we can not capture the high-order and complex interactions among parts~\cite{chen2019mixed}.
Therefore, in our data-free framework, we empirically exploit the spatial attention for our generator and the mixed high-order attention for distillation.

\subsection{Data-free Distillation}
Data-free Distillation has become a hot topic in recent years, mainly due to privacy protection~\cite{liu2021data}. 
It exploits synthesized alternative samples to solve the dilemma that model can not directly access the original data and makes gratifying achievements in the task of classification~\cite{yin2020dreaming,chen2019data,fang2019data,fang2021contrastive}. For example, ADI~\cite{yin2020dreaming} utilizes batch normalization statistics~(BNS) of the pre-trained teacher to optimize the noise to synthesize high-fidelity images for KD.
CMI~\cite{fang2021contrastive} exploits the local and global contrast of samples to optimize the generator diversity.
This kind of method ordinarily can synthesize more realistic images and achieve relatively better performance.
DFAL~\cite{chen2019data} adopts a generator to synthesize images, and then the student learns the knowledge from the teacher by distillation. ZSKT~\cite{micaelli2019zero} exploits the adversarial distillation to transfer the knowledge from teacher to student by KL and spatial attention, while DFAD~\cite{fang2019data} only utilizes the MAE loss to fit the output distribution of the teacher.
All kinds of the above methods can achieve relatively inspiring achievements in coarse-grained classification, and there is no specific research on FGVC.
Motivated by this, we conduct the study for data-free fine-grained distillation.

\section{Preliminary}
\label{sec:preliminary}
Our approach follows the basic thinking of DFKD, as depicted in Fig.~\ref{fig:framework}. First, a generator $\mathcal{G}$ is employed to synthesize a batch of images from noise $z \sim \mathcal{N}(0,1)$, $\mathcal{G}(z) \to \hat{x}$, $\mathcal{B} = \left \{\hat{x}_{1},\hat{x}_{2},..., \hat{x}_{n} \right \}, n \in \{1,..., \mathrm{N}\}$, where $\mathrm{N}$ is the batch size. Then the synthesized image $\hat{x}$ is input to the pre-trained teacher $\mathcal{T}$ and student $\mathcal{S}$ to support their distillation. Finally, the generator $\mathcal{G}$ is optimized by adversarial distillation. \\
\textbf{Data-free Adversarial Distillation.}
Essentially, Data-free adversarial distillation is a robust minimax optimization problem~\cite{ben2009robust}, which encourages the generator to minimize the possible loss for a worst-case scenario (maximum loss) through adversarial training under data uncertainty. In the data-free scenario, it can be denoted as
\begin{equation}
\label{eq:DFAD}
\min_{\mathcal{S}}\max_{\mathcal{G}} \left \{\mathbb{E}_{p(z)} \left [\mathcal{D}(\mathcal{T}(\mathcal{G}(z)), \mathcal{S}(\mathcal{G}(z))) \right ]  - \delta \mathcal{L}_{\mathcal{G}}  \right \} \,,
\end{equation}

where $\mathcal{D}$ represents the discrepancy measure, which normally exploits the Kullback-Leibler (KL) divergence as an optimization term.
$\delta \ge 0$ is the balance factor, and $\mathcal{L}_{\mathcal{G}}$ is the optimization term of generator $\mathcal{G}$. \\
\textbf{Knowledge Distillation}. According to the principle of classic knowledge distillation~\cite{hinton2015distilling}, the soft output of the network (a.k.a.~probability distribution) implies the similarity between the current sample and other categories.
Therefore, traditional methods~\cite{choi2020data,domomentum,nayak2019zero} usually adopt the KL Divergence to measure the difference between the two distributions of teacher and student.
The probability distribution distillation can be formulated as
\begin{equation}
\label{eq:KD}
\mathcal{L}_{\mathrm{KD}}  = \mathbb{E}_{\hat{x}}\left[D_{\mathrm{KL}}\left(\sigma(\mathcal{S}(\hat{x})) \| \sigma(\mathcal{T}(\hat{x})) \right) \right] \,,
\end{equation}
where $D_{\mathrm{KL}}$ represents the Kullback-Leibler (KL) divergence, and $\sigma$ is the softmax operation.

\noindent\textbf{Prior Information Regularization.} Prior information regularization aims to regularize the feature distribution of synthesized images by prior distribution information, i.e., mean $\mu$ and variance $\sigma^2$ of BatchNorm~\cite{yin2020dreaming}, which motivates the synthetic samples to approach the distribution of the original samples.
{\small
\begin{equation}
\label{eq:BN}
\mathcal{L}_\mathrm{BN} = \min_{\mathcal{G}} \sum_{l} \left \| \mu_{l} - \mu_{l}(\mathcal{G}(z)) \right \|_{2}
                 + \left \| \sigma_{l}^{2} - \sigma_{l}^{2}(\mathcal{G}(z)) \right \|_{2} \,,
\end{equation}
}
where $l$ donates the $l^{th}$ BatchNorm layer of the teacher model, $\mu$ and $\sigma^2$ are the batch-wise mean and variance, respectively.

\begin{figure}[!t]
  \centering
  \includegraphics[width=\linewidth]{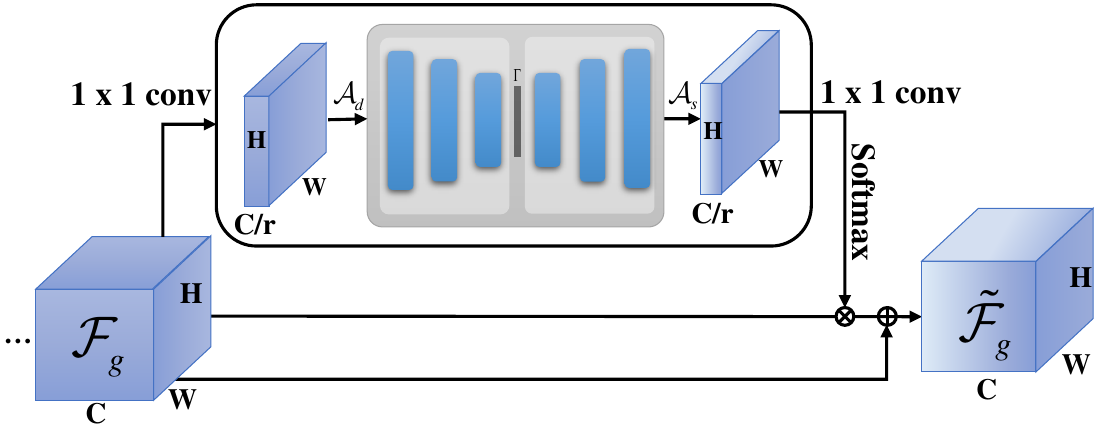}
  \vspace{-3mm}
  \caption{The spatial attention module of the generator, in which $\otimes $ denotes the element-wise multiplication and $\oplus$ denotes the element-wise addition.}
  \label{fig:gam}
  \vspace{-2mm}
\end{figure}

\section{Proposed Approach}
\subsection{Discriminative Feature Synthesis}
In the DFKD framework, it is common to exploit a generator to assist in generating alternative samples for coarse-grained classification. 
However, directly applying it to synthesize fine-grained samples often does not yield desirable discriminative features. 
This is because the conventional generator cannot focus on subtle discriminative features, which decreases the ability of teacher to extract representation from various semantic parts and thus hampers the effectiveness of the distillation. 
Differing from the traditional approaches, in our framework, we employ a DCGAN~\cite{radford2015unsupervised,teoh2022deep} generator with the attention module to increase the representation ability of features and tell the generator where to focus.
Inspired by preceding attention works such as CBAM~\cite{woo2018cbam} and CBM~\cite{parkbam}, which stacks channel attention and spatial attention in series, we exploit the attention mechanism in our approach. 
However, unlike the prior approaches, we implement the attention by the encoder-decoder manner, thinking that the non-linear convolution can pay attention to context knowledge of features, which is more suitable for dense generation tasks. Besides, in order to have stability training, the spectral normalization~\cite{miyatospectral} is exploited to regularize the ConvTranspose2d layers of DCGAN, which controls the weights of modules by the Lipschitz constant.  

Concretely, as displayed in Fig.~\ref{fig:gam}, the noise $z$ is input to generator $\mathcal{G}$ to synthesize the alternative samples $\hat{x}$. We first divide the whole DCGAN module into four blocks.
Then we plug the attention module at each block to compute the low-dimensional feature maps $\mathcal{A}_{d} \in \mathbb{R}^{\mathrm{C}/r \times \mathrm{H} \times \mathrm{W}}$ from original feature maps $\mathcal{F}_{g} \in \mathbb{R}^{\mathrm{C} \times \mathrm{H} \times \mathrm{W}}$, where  $r$ is the scaled scalar, $\mathrm{C}$ denotes channel, $\mathrm{H}$ and $\mathrm{W}$ represent the size of the feature maps. This aims to achieve lightweight feature maps.
Next, the encoder is employed to achieve the latent space as follows:
\begin{eqnarray}
  \label{eq:channel}
  \left\{
  \begin{aligned} 
    \mathcal{A}_{d} &= \mathrm{Cov}^{1 \times 1}(\mathcal{F}_{g}) \,, \\ 
    \Psi  & =  \mathrm{ReLU}(\mathrm{BN}(\mathrm{Cov}^{3 \times 3}(\mathcal{A}_{d}))) \,, \\ 
    \Gamma & =  \mathrm{ReLU}(\mathrm{BN}(\mathrm{Cov}^{3 \times 3}(\mathrm{MP}(\Psi)))) \,,
  \end{aligned}
  \right.
\end{eqnarray}
where $\Psi$ represents features of intermediate process,  $\mathrm{Cov}^{1 \times 1}$ and $\mathrm{Cov}^{3 \times 3}$ denote the convolution with kernel size of $1 \times 1$  and $3 \times 3$, and MP represents maximum pooling.

By Eq.~\ref{eq:channel}, we can get the representation of low-dimensional latent space $\Gamma$ from $\mathcal{A}_{d} \in \mathbb{R}^{\mathrm{C}/r \times \mathrm{H} \times \mathrm{W}}$,
and then decode the space with maximum uppooling~(MUP) to achieve spatial-wise attention $\mathcal{A}_{s}  \in \mathbb{R}^{\mathrm{C}/r \times \mathrm{H} \times \mathrm{W}}$.
This operation can preserve information of the key locations in the feature to achieve the 2D spatial attention map $\mathcal{A}_{s}$ as follows: 
\begin{eqnarray}
  \label{eq:spatial}
  \left\{
  \begin{aligned}
    \Psi  & = \mathrm{MUP}(\mathrm{ReLU} (\mathrm{BN}(\mathrm{DC}^{3 \times 3}(\Gamma)))) \,, \\ 
    \mathcal{A}_{s} & = \mathrm{Cov}^{1 \times 1}(\mathrm{ReLU}(\mathrm{BN}(\mathrm{DC}^{3 \times 3}(\Psi)) )) \,,
  \end{aligned}
  \right.
\end{eqnarray}
where $\mathrm{DC}^{3 \times 3}$ denotes the deconvolution with kernel size of $3 \times 3$, MUP represents the maximize unpooling. 
Then, aggregating the attention maps to the original feature maps to achieve $\mathcal{\tilde{F}}_{g}$ is formulated as: 
\begin{equation}
  \label{eq:sim}
  \mathcal{\tilde{F}}_{g} = \lambda( \mathrm{Softmax}(\mathcal{A}_{s}) \times \mathcal{F}_{g}) + \mathcal{F}_{g}  \,,
  \end{equation}
where $\lambda$ is the hyperparameter to balance the attention maps with features, which defaults to $5e^{-2}$ in our experiments. More details about the contributions of the attention generator are presented in Tab.~\ref{tab:generator}. 

\subsection{Mixed High-Order Attention Distillation}
\begin{figure}[!t]
  \centering
  \includegraphics[width=\linewidth]{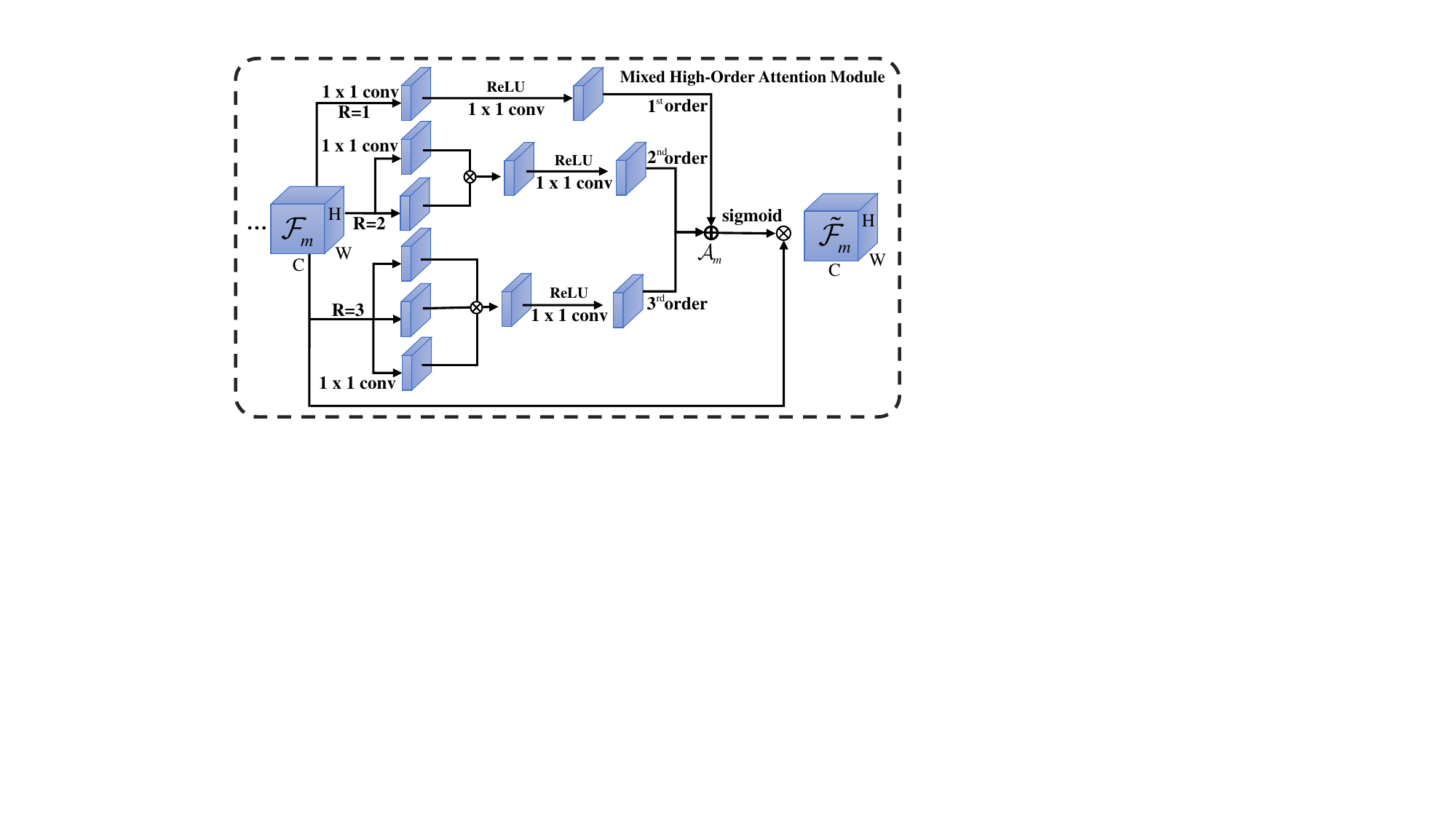}
  \vspace{-2mm}
  \caption{The MHA module of teacher and student in distillation stage. }
  \label{fig:mhad}
  \vspace{-1.5em}
\end{figure}
In the stage of distillation, traditional DFKD methods~\cite{chen2019data,fang2019data,micaelli2019zero} to solve coarse-grained classification commonly exploit the distribution of output layers due to the significant inter-class variation (compared to intra-class variation), which enables deep networks to learn generalized discriminatory features of coarse-grained classification.
However, the distribution knowledge distillation only exploits category-related information with dark knowledge~\cite{hinton2015distilling}, which lacks semantically relevant information. We argue that this paradigm may not be ideal for FGVC, due to the data-free scenario.

To solve the above difficulties, recent methods commonly exploit attention mechanism~\cite{cai2017higher, min2020multi} to capture the discriminative features of the object. 
However, the existing FGVC methods of attention mechanism are mainly designed for data-available scenarios, and there is no related research in the data-free scenario. 
This motivates us to extend this strategy in a data-free setting.
Besides, the related attention distillation works~\cite{zhou2020channel, shu2021channel} only consider the low-order attention information, which only focuses on the local information and cannot capture the complex interactions among parts, resulting in less discriminative attention proposals and failing in capturing the subtle differences among objects.
In the data-free scenario, due to the semantic information being sparse~\cite{ding2019selective}, we believe that low-order attention distillation cannot fully express the knowledge of the features.
Thus we propose to exploit mixed high-order attention~(MHA) to distill the aggregated local features and semantic context relation of synthesized FGVC images.

Our mixed high-order attention module is shown in Fig.~\ref{fig:mhad}, in which mixed 3-order attention~(i.e., $\mathrm{R} = 3$) is exploited. The feature $\mathcal{F}_{m} \in \mathcal{R}^{\mathrm{H} \times \mathrm{W} \times \mathrm{C}}$ is first extracted by three route $1 \times 1$ convolutions to achieve 3-order intermediate representations. In each route, the convolution layer and produced relative representation are the same as the order $\mathrm{R}$. Then we multiply the representations of each order to obtain aggregated representations. For each aggregated representation, we exploit RELU and $1 \times 1$ convolution to produce the new map which will be aggregated with global attention maps $\mathcal{A}_{m}$.  At last, the activated global attention map $\mathcal{A}_{m}$ will be multiplied with the original features $\mathcal{F}_{m}$ to produce the final attention maps $\mathcal{\tilde{F}}_{m} =  \mathcal{A}_{m} \times   \mathcal{F}_{m}$. 

For teacher and student, their channels may be different.
Thus we first exploit the $Adapter$ to upgrade the channel of the student to the same number as the teacher, which is also implemented by the $1 \times 1$ 2D convolution. Therefore, at each block of the intermediate layer of $\mathcal{T}$ and $\mathcal{S}$, we exploit mean square error~(MSE) to measure the MHAD loss, which is formulated as:
 \begin{eqnarray}
  \mathcal{L}_{\mathrm{MHAD}} = \frac{1}{\mathrm{N} \times \mathrm{C}} \sum_{i=1}^{\mathrm{N}}\sum_{j=1}^{\mathrm{C}} \mathrm{MSE}( \mathcal{F}^{t}_{m}, \mathcal{F}^{s}_{m} ) \,,
\end{eqnarray}
where $\mathrm{N}$ and $\mathrm{C}$ represent the batch size and channel, while $\mathcal{F}^{t}_{m}$ and $\mathcal{F}^{s}_{m}$ denote the attention map of an intermediate block of teacher and student, respectively. It should be noted that this strategy is only exploited during our training process, which does not participate in the inference. Therefore, this does not affect the efficiency of the model.

\begin{algorithm}[!t]
  \caption{The whole pipeline of DFKD-FGVC.}
  \label{alg:pipeline}
  \textbf{Input}: A pre-trained teacher model $\mathcal{T}$ on real data, generator $\mathcal{G}$ and student network $\mathcal{S}$. \\
  \textbf{Output}: A well-trained student network $\mathcal{S}$.

  \begin{algorithmic}[1]
    \STATE \textbf{// Ganerator Stage} \\
    \FOR{\textit{number of iterations}}
    \FOR{\textit{t steps iterations}}
    \STATE Generate random noise $z \sim \mathcal{N}(0,1)$ ; \\
    \STATE Synthesize supporting sample $\hat{x} = \mathcal{G}(z)$ ; \\

    \STATE Optimize the generator by $\mathcal{L}_{\mathrm{BN}}$, and $-\mathcal{L}_{\mathrm{KD}}$; \\
    \STATE Freeze $\mathcal{S}$ and $\mathcal{T}$, and update $\mathcal{G}$ by Eq.~\ref{eq:gen} . \\
    \ENDFOR
    \ENDFOR

  \STATE \textbf{// Distillation Stage}\\
  \FOR{\textit{number of iterations}}
  \FOR{\textit{k steps iterations}}
  \STATE Generate random noise $z \sim \mathcal{N}(0,1)$ ; \\
  \STATE Synthesize supporting sample $\hat{x} = \mathcal{G}(z)$ ; \\
  \STATE Calculate discrepancy by $\mathcal{L}_{\mathrm{KD}}$, $\mathcal{L}_{\mathrm{MHAD}}$, and $\mathcal{L}_{\mathrm{SFCL}}$. 
  \STATE Freeze $\mathcal{G}$ and $\mathcal{T}$, and update $\mathcal{S}$ by Eq.~\ref{eq:stu} ; \\
  \ENDFOR
  \ENDFOR
  \end{algorithmic}
\end{algorithm}

\renewcommand\arraystretch{1.1} %
\newcolumntype{Y}{>{\centering\arraybackslash}X}
\begin{table*}[!ht]
  \centering
  \vspace{-3mm}
  \caption{Results of different data-free distillation methods on three fine-grained datasets.}
  \vspace{2mm}
   \resizebox{0.8\linewidth}{!}{
    \begin{tabular}{c|c|c|c|c|cccc}
    \Xhline{1.05pt}
    & Setting & Prior Info. & \multicolumn{2}{c|}{Compression Info.} & \multicolumn{3}{c}{Accuracy} \\
    \hline
    Method &  \multicolumn{1}{c|}{Data-free}& \multicolumn{1}{c|}{BN} & \multicolumn{1}{c|}{FLOPs} &\multicolumn{1}{c|}{Params.} & Aircraft &  Cars196 & CUB200 \\
    \hline
    ResNet-34~(T.) & $\times$  & $\times$  & $\sim$3.67G   &  $\sim$22M  & 70.15    & 84.22 &   76.87 \\
    ResNet-18~(S.) & $\times$  &  $\times$ & $\sim$1.82G  &  $\sim$11M  &  68.71 &     77.43 &      58.60 \\
    \hline    
    ZSKD~\cite{nayak2019zero} & $\checkmark$ & $\times$   & $\sim$1.82G  &   $\sim$11M  & 37.32 &       26.21    &   30.53   \\  
    ZSKT~\cite{micaelli2019zero} & $\checkmark$ & $\times$   & $\sim$1.82G  &   $\sim$11M  & 51.16 &       28.48    &   31.88   \\   
    DAFL~\cite{chen2019data} & $\checkmark$ & $\times$   & $\sim$1.82G  &   $\sim$11M  & 43.69    & 37.71 &     31.01 \\
    DFAD~\cite{fang2019data} & $\checkmark$ & $\times$   & $\sim$1.82G  &   $\sim$11M  & 49.51 &  48.72 &      40.15 \\
    ADI~\cite{yin2020dreaming} & $\checkmark$ & $\checkmark$   & $\sim$1.82G  &   $\sim$11M  & 58.14    & 65.24 &  47.63 \\
    DFQ~\cite{choi2020data} & $\checkmark$ & $\checkmark$   & $\sim$1.82G  & $\sim$11M
    & 60.22   & 66.14 &  48.43   \\
    MAD~\cite{domomentum} &$\checkmark$ & $\checkmark$   & $\sim$1.82G  & $\sim$11M  & 63.74    & 67.53 &  53.43   \\
    CMI~\cite{fang2021contrastive}& $\checkmark$ & $\checkmark$   & $\sim$1.82G  &   $\sim$11M  & 63.57 &  68.74 &     53.53 \\
    \hline
    \textbf{Ours} & $\checkmark$ & $\checkmark$  & $\sim$1.82G  &   $\sim$11M & \textbf{65.76} &  \textbf{71.89} & \textbf{56.93} \\
    \Xhline{1.05pt}
    \end{tabular}%
  }
  \vspace{-3mm}
  \label{tab:main}%
\end{table*}%

\subsection{Semantic Feature Contrast Learning}
Since the pre-trained teacher has a higher discriminative ability than the student, optimizing the student by comparing the features of the teacher is conducive to improving the ability of the student to distinguish right from wrong.
Therefore, in our FGVC task, we not only focus on intermediate low-level feature variances but also high-level semantic variances of the penultimate layer. Unlike traditional paradigms~\cite{chen2021wasserstein,chen2020simple,khosla2020supervised,tian2019contrastive}, which contrast the original~\cite{chen2020simple,khosla2020supervised} and augmentation data or in data-driven scenarios~\cite{chen2021wasserstein,tian2019contrastive}.
we exploit high-level semantic features to contrast feature representation of teacher and student and aim to learn the variances between different categories in the data-free scenarios, which are more difficult than data-driven scenarios.

Specifically, in the penultimate layer, we obtain their semantic feature representations and exploit the multi-layer perceptron~(MLP)  to map the representations to a common space to achieve $2\mathrm{N}$ feature representations as $\mathcal{F}_{s} = \mathcal{C}(\mathcal{S}(\mathcal{G}(z)))$ and $\mathcal{F}_{t} = \mathcal{C}(\mathcal{T}(\mathcal{G}(z)))$, where $\mathcal{C}$ is the MLP layer with two hidden linear layers.
Then, we normalize the features to a unit hyperspace and measure their similarity as follows:
\begin{equation}
\label{eq:sim}
sim(\mathcal{F}_{t}, \mathcal{F}_{s}) =  \frac{\mathcal{F}_{t} \cdot \mathcal{F}_{s}^{\top}}{\left \| \mathcal{F}_{t} \right \| \cdot \left \| \mathcal{F}_{s} \right \| } \,,
\end{equation}
where $\cdot $ denotes the inner (dot) product.
The cosine distance is used as the similarity metric to measure the relationship between two feature representations for contrastive loss, which is defined as
{\small
\begin{equation}
  \label{eq:SFRC}
  \mathcal{L}_\mathrm{SFCL}=  \min_{\mathcal{S}} \left \{ - \log \frac{\exp (sim (\mathcal{F}_{t}^{i}, \mathcal{F}_{s}^{j}) / \tau)}{\sum_{k}^{2\mathrm{N}} \mathbbm{1}_{[k \neq i]} \exp (sim (\mathcal{F}_{t}^{i}, \mathcal{F}_{s}^{k} ) / \tau)} \right \} \,,
\end{equation}
}
where $\mathbbm{1}_{[k \neq i]}$ is an indicator function that returns 1 if $i = j$,  $i$ and $j \in 2\mathrm{N}$  are the indexes of the samples in the representations, and $\tau$ denotes a temperature parameter. This loss maximizes the representations of the different categories, where the teacher can extract the effective features from noisy images and pull away from the other dissimilar features. Therefore, if one feature of the teacher is viewed as an {\it anchor}, and the student extracts another representation of this synthetic image as the {\it positive}. Due to the weak ability of sample representations of the student model, such operation of the student plays a role as augmented images.
The other $2(\mathrm{N}-1)$ features can be viewed as the {\it negative}. 
Therefore, the loss $\mathcal{L}_{\mathrm{SFCL}}$ is used to optimize the student to close to the teacher model, i.e., improving the ability of students to distinguish different samples.
\subsection{Total Objects}
In the whole algorithm pipeline~\ref{alg:pipeline}, we first optimize the generator to synthesize more realistic diverse samples. The total objective of the generator is
\begin{equation}
  \label{eq:gen}
\min_{\mathcal{G}}
\alpha \mathcal{L}_{\mathrm{BN}} - \mathcal{L}_{\mathrm{KD}}  \,.
\end{equation}
Then, with the above strategy for the generator, we can detail the total objective of the student:
\begin{equation}
  \label{eq:stu}
\min_{\mathcal{S}} \mathcal{L}_{\mathrm{KD}}+\beta\mathcal{L}_{\mathrm{MHAD}}+\gamma  \mathcal{L}_{\mathrm{SFCL}} \,,
\end{equation}
where $\alpha$, $\beta$, and $\gamma $ are both hyper-parameters. The training plays an adversarial distillation to optimize both at each iteration.

\section{Experiments}
\subsection{Datasets and Implementation Details}
\noindent\textbf{Datasets.} To demonstrate the effectiveness of our approach, we conduct experiments on three fine-grained datasets. \\
\textbf{Aircraft.}~FGVC-Aircraft~\cite{maji2013fine} contains 100 different aircraft variants formed by 10,000 annotated images, which is divided into two subsets, i.e., the training set with 6,667 images and the testing set with 3,333 images. \\
\textbf{Cars196.}~The Stanford Cars dataset~\cite{krause20133d} contains 16,185 images from 196 categories of cars. The data is split into 8,144 training images and 8,041 testing images. \\
\textbf{CUB200.}~The Caltech-UCSD birds dataset (CUB-200-2011)~\cite{welinder2010caltech} consists of 11,788 annotated images in 200 subordinate categories, including 5,994 images for training and 5,794 images for testing. \\
\noindent\textbf{Implementation Details.} Our method is implemented with the PyTorch library. All the models are trained on NVIDIA 3090 GPUs with 24G memory.
\begin{figure*}[!htbp]
  \centering
  \includegraphics[width=\linewidth]{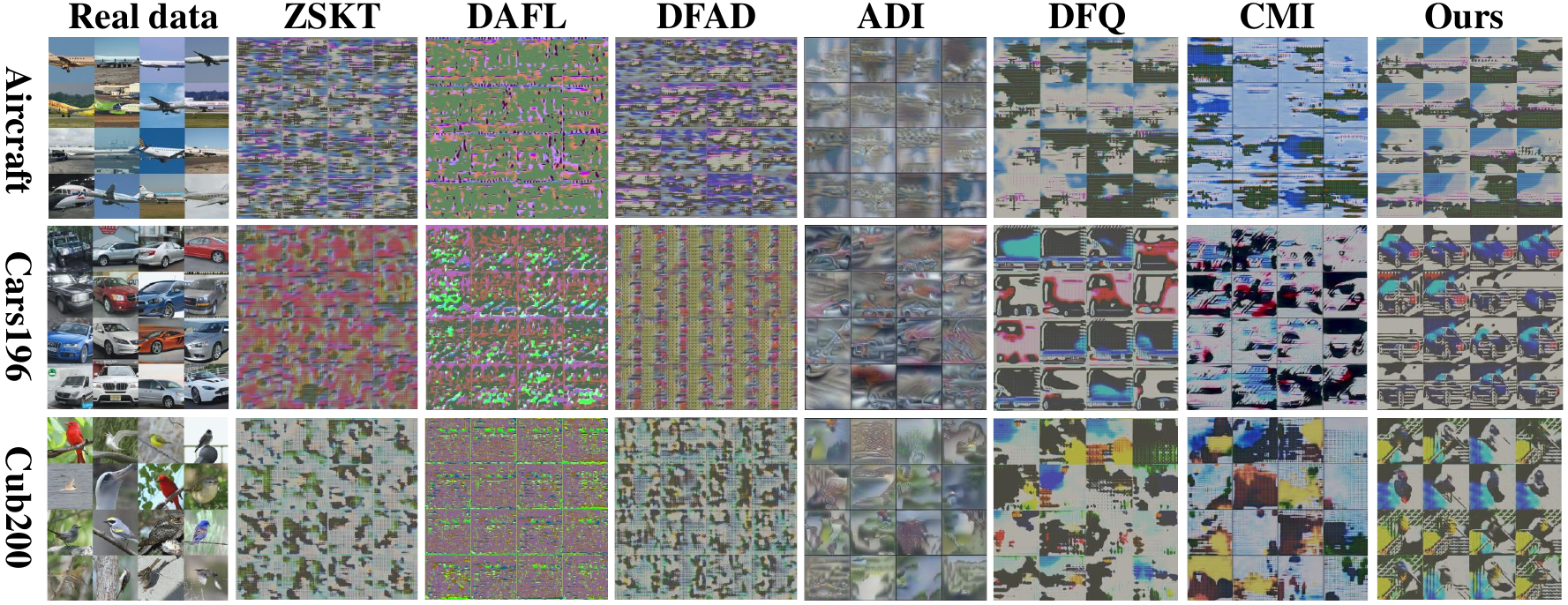}
  \caption{Visualization synthetic images generated by some representative approaches on Aircraft, Cars196, and CUB200 datasets.}
  \vspace{-4mm}
  \label{fig:gen}
\end{figure*}
ResNet-34~\cite{he2016deep} is employed as the cumbersome teacher network for all experiments in this paper, and four architectures, i.e., ResNet-18~\cite{he2016deep}, WRN40-2~\cite{he2016deep}, MobileNetV2~\cite{sandler2018mobilenetv2}, and ResNet-34~\cite{he2016deep} are utilized as students.
We first train the generator for 20 steps~(i.e., $t$=20) where the generator follows the architecture of DCGAN~\cite{radford2015unsupervised}.
Adam~\cite{kingma2014adam} is adopted to optimize the generator with an initial learning rate of $1\times 10^{-3}$ and $beta$ is set 0.5 to 0.99.
Then, we train the student 15 steps~(i.e., $k$=15) after the generator and optimize the parameters by the SGD optimizer with a momentum of 0.9, a batch size of 64 as default, and a weight decay of 5$\times 10^{-4}$.
The initial learning rate starts at 1$\times 10^{-2}$ with cosine annealing for a total of 200 epochs.
In the pre-trained stage, due to subtle discrepancies that are difficult to detect, the input images of fine-grained datasets are both resized and randomly cropped to 224$\times$244.
In the data-free distillation stage, all the synthetic images are the same as the size of the original input images in the pre-trained stage.
As for the hyper-parameters, both $\alpha$, $\beta$, and $\gamma$ are set to 0.3, 10, and 8 by default, respectively.
Floating point operations (FLOPs) and parameters (Params) are employed to measure the computation and storage cost of the networks.
\subsection{Results and Comparisons}
As shown in Table~\ref{tab:main}, we focus on evaluating our approach and other compared methods on three public fine-grained datasets, i.e., Aircraft, Cars196, and Cub200. To evaluate the effectiveness of our proposed method, we conduct fair comparison experiments with two kinds of DFKD methods which are primarily for general classification tasks: (1) Without~($\times$) prior information methods, including ZSKT, DAFL, and DFAD; (2) With~($\checkmark$) prior information methods, including ADI, DFQ, MAD, and CMI.
The first two rows of the table show the results of the teacher and student with annotated data supervision in training, which is also our target to achieve by KD.
Obviously, the performance of the methods exploiting prior information is better than those without. 
For example, DFAD only achieves 49.51\%, 48.72\%, and 40.15\% on three datasets, while ADI can achieve 58.14\%, 65.24\%, and 47.63\%.
This is mainly because BN regularization has a good performance to inverse and generate relatively realistic images, which is particularly important for downstream distillation.
Based on the BN regularization, our approach exploits the spatial attention generator to generate the images with semantic information, which can further improve the performance of the student.

Besides, almost all of the above approaches exploit the vanilla KD~(e.g., KL divergence) to transfer the knowledge from the output layer, although they can perform well on coarse-grained classification, but do not perform well on fine-grained classification.
Our method mainly adopts two strategies to further improve the performance of the student by about 3\% on average, which indicates that vanilla KD alone cannot complete all knowledge transfer, and special design is necessary for FGVC tasks distillation in DFKD.
Under identical conditions, thanks to two optimization strategies, i.e., MHAD and SFCL, our approach outperforms the other data-free methods to achieve state-of-the-art performance on three datasets. 

\begin{table}[!htbp]
  \centering
  \caption{More comparisons of different architectures' students with ResNet-34 on Aircraft dataset.}
  \vspace{2mm}
  \resizebox{\linewidth}{!}{
    \begin{tabular}{c c c c c c c c c c}
    \toprule
     Student & ZSKT & DFAL & DFAD & ADI & DFQ & MAD & CMI & \textbf{Ours}  \\
    \hline
    WRN40-2 & 49.13    &  36.83  &  50.44  &  57.83& 58.26 & 59.85 & 62.43 & \textbf{64.54}  \\
    MobileNetV2 & 24.39 & 18.51 & 23.01  &   53.66 &53.93 &54.61  & 55.04 & \textbf{57.37} \\
    ResNet-34 &39.52 & 36.63 &52.15    &  60.75 & 61.75 & 63.12 & 64.66 &\textbf{65.48} \\
    \bottomrule
    \end{tabular}%
  }
  \vspace{-3mm}
  \label{tab:archresult}%
\end{table}%

To verify the generality of our approach, we perform distillation on another three student models with different architectures, including heterogeneous distillation~(i.e., WRN40-2 and MobileNetV2) and self-distillation~(i.e., Resnet-34). 
For WRN40-2 and MobileNetV2, we leverage the MLP with two hidden layers to map the dimension to match the teacher and implement our two strategies both in the penultimate layer.
As shown in Table~\ref{tab:archresult}, our approach can also achieve state-of-the-art performance in different architectures.

\begin{figure*}[!htbp]
  \vspace{-2mm}
  \hspace{-3mm}
  \centering
  \subfigure[ZSKT]{
    \includegraphics[width=0.16\linewidth]{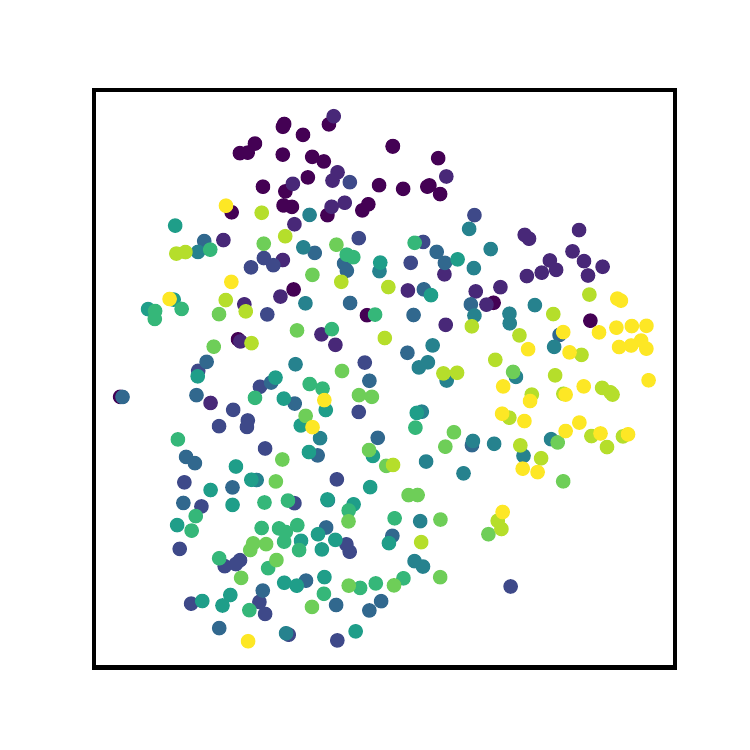}
    \label{fig:tsezskd}
  }
  \hspace{-3mm}
  \subfigure[DFAD]{
    \includegraphics[width=0.16\linewidth]{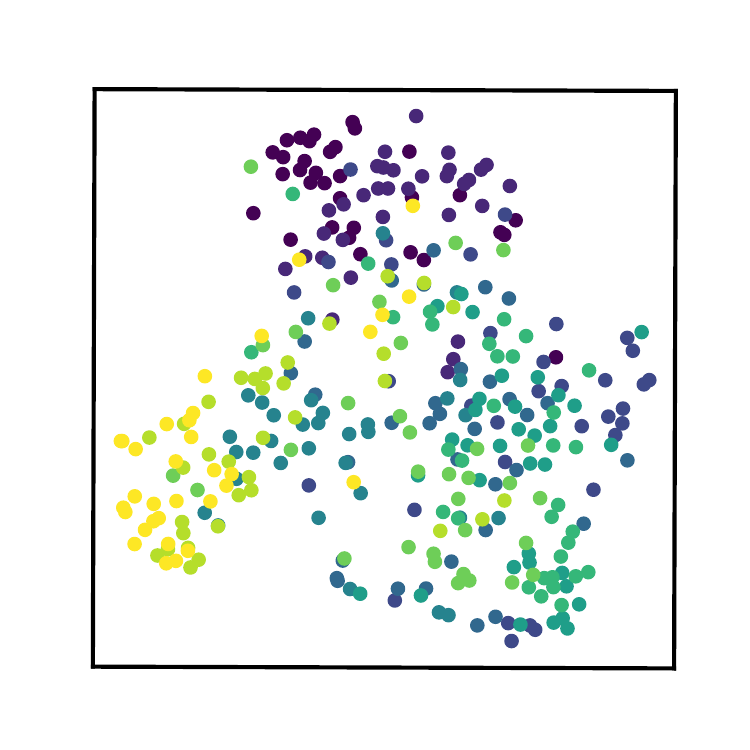}
    \label{fig:tsedfad}
  }
  \hspace{-3mm}
  \subfigure[ADI]{
    \includegraphics[width=0.16\linewidth]{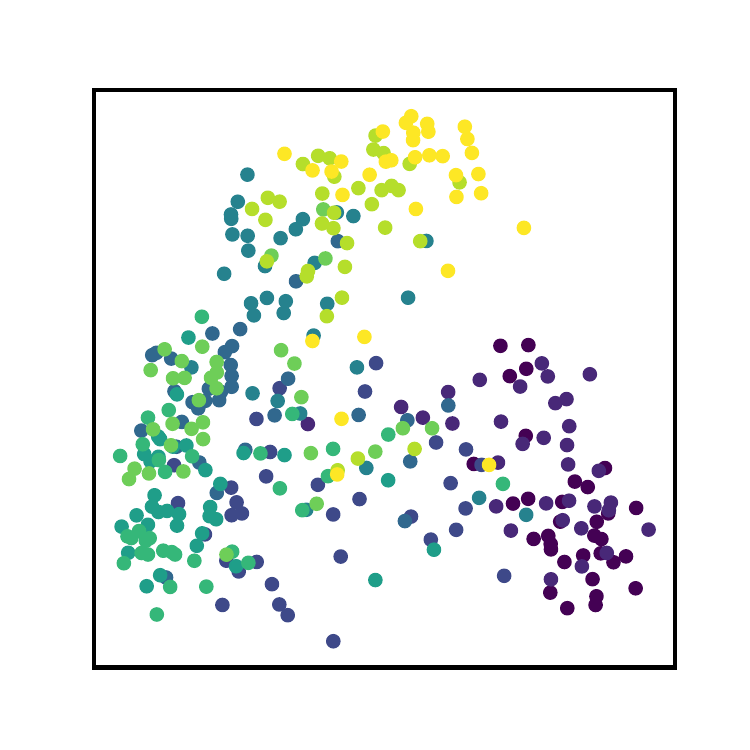}
    \label{fig:tsedadi}
  }
  \hspace{-3mm}
  \subfigure[DFQ]{
    \includegraphics[width=0.16\linewidth]{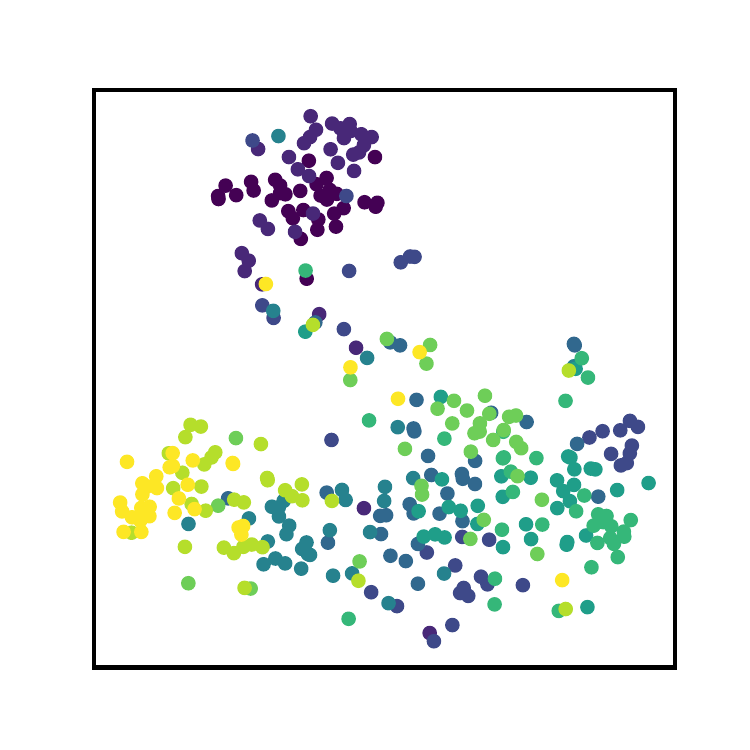}
    \label{fig:tsedcmi}
  }
  \hspace{-3mm}
  \subfigure[CMI]{
    \includegraphics[width=0.16\linewidth]{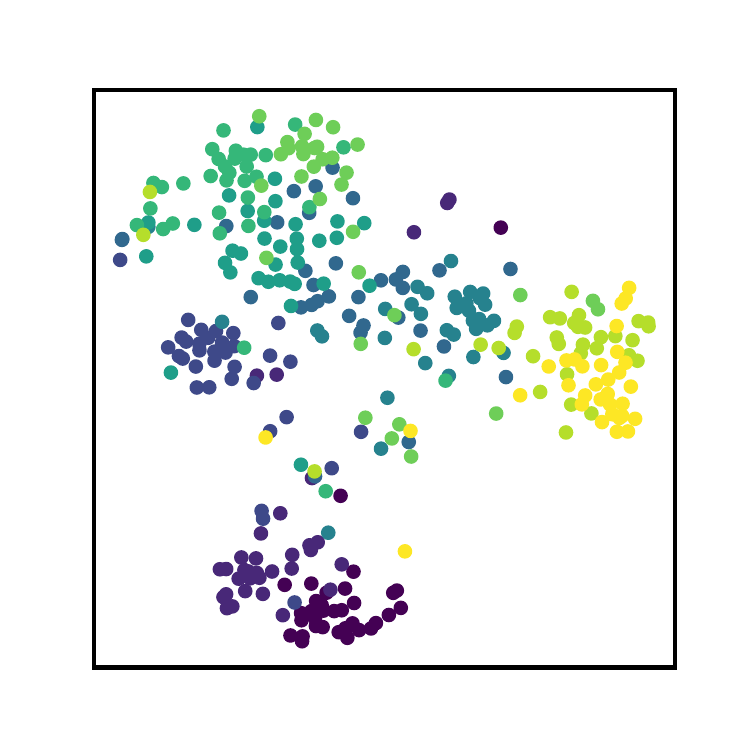}
    \label{fig:tseus}
  }
  \hspace{-3mm}
  \subfigure[Ours]{
    \includegraphics[width=0.16\linewidth]{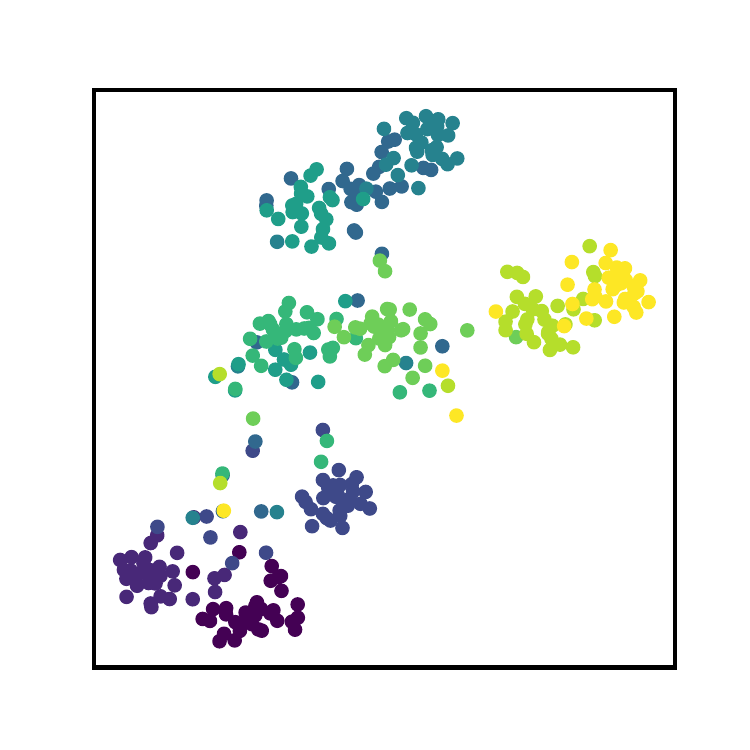}
    \label{fig:tsemobile}
  }
  \caption{Visualization of t-SNE distribution on Aircraft dataset. }
  \label{fig:tsne}
  \vspace{-4mm}
\end{figure*}

\subsection{Visualization and Analysis}
\label{sec:vis}
\noindent\textbf{Synthetic images.} To clearly evaluate the effect of synthesized images, we present a visualization analysis of some representative methods on Aircraft, Cars196, and CUB200 in this section.
As we can see from Fig.~\ref{fig:gen}, the first column is the real data for reference. However, for ZSKT, DAFL, and DFAD, there is a big gap between the generated images and the real data.
Since ADI, CMI, and Ours both exploit the BN to regularize the features, the synthesized images are more realistic than the other data-free methods, which is beneficial for downstream distillation.
With the assistance of two optimization strategies of MHAD, and SFCL, our approach can generate better and more discriminative foreground images compared to ADI, DFQ, and CMI.
For example, we can clearly distinguish the outline of the car and the color of the different areas of the birds.

\noindent\textbf{t-SNE.} To illustrate the advantages of our approach in synthesizing images having more similar distributions with real images, we sample 10 categories from the Aircraft dataset and visualize the representations of MobileNetV2 by t-SNE as Fig.~\ref{fig:tsne}.
As shown in~Fig.~\ref{fig:tsemobile}, our approach gains obviously better representations than the other methods, according to the comparison with each other.
Compared the Tab.~\ref{tab:main} with Fig.~\ref{fig:gen}, we can conclude that the performance of the student primarily relies on the quality of the synthetic images and the effect of knowledge transfer in DFKD.

\begin{figure}[!ht]
  \centering
  \includegraphics[width=\linewidth]{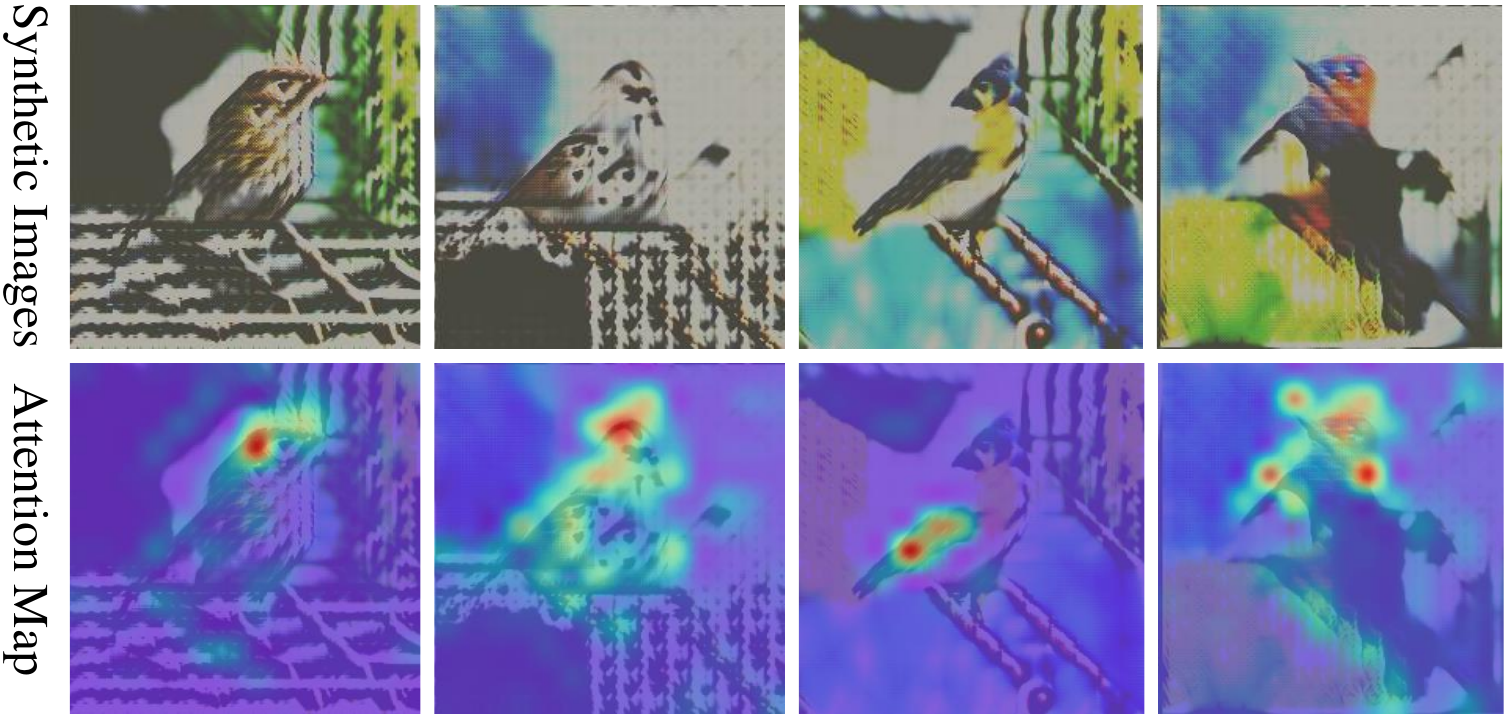}
  \caption{Visualization of synthetic images with attention map generated by GradCAM on CUB200 datasets.}
  \vspace{-4mm}
  \label{fig:attention}
\end{figure}

\noindent\textbf{Attention map.} To further verify the effect of our mixed high-order attention~(MHA) modules feature selection, we visualized the generated samples through GradCAM~\footnote{https://github.com/jacobgil/pytorch-grad-cam.git}, as shown in Fig.~\ref{fig:attention}.
The first row is the synthesized alternative samples of CUB200 which are generated by our attention module. We can see the fine-grained semantic information of different synthesized birds. For example, we can distinguish different beaks or wings of birds, and different colors of features. When we employ the student embedded with MHA modules to visualize birds' discriminative features by GradCAM, the attention maps are sparse and focus on the discriminative parts, as shown in the second row of the figure. For example, the wings of birds are activated, which indicates that the wings are being paid attention to. 
We can conclude that MHA modules can focus on contextual semantic information on features which is based on the attention of discriminative features.

\subsection{Ablation Study}
\noindent\textbf{Contribution of loss.} 
To verify the contribution of each component, we conduct ablation experiments on the three datasets with ResNet-18, as shown in Table~\ref{tab:ablation_study}.
In the first row is the Baseline of each benchmark, which exploits the Eq.~\ref{eq:gen} to optimize the $\mathcal{G}$, while only optimizing the $\mathcal{S}$ by exploiting the $\mathcal{L}_{\mathrm{KL}}$ to distill the knowledge.
Then, adding the $\mathcal{L}_{\mathrm{SFCL}}$  component to the Baseline, the result of each benchmark is improved by 3.07\%, 2.33\%, 2.92\%, respectively. Likewise, when we add $\mathcal{L}_{\mathrm{MHAD}}$  to the baseline, it can also achieve significant improvement. Nevertheless, by comparing both, we can find that the contribution of $\mathcal{L}_{\mathrm{SFCL}}$ is relatively weaker than $\mathcal{L}_{\mathrm{MHAD}}$, which proves the effectiveness of exploiting mixed high-order attention to model discriminative features, which has been ignored by other methods. Finally, we add both to the baseline and obtain the final state-of-the-art effect.
\begin{table}[!htbp]
  \centering
  \caption{The ablation study of our approaches with different components. `+' denotes the add operation.}
  \vspace{2mm}
  \resizebox{0.9\linewidth}{!}{
    \begin{tabular}{lccc}
    \toprule
    Method &  Aircraft & Cars196 & Cub200  \\
    \hline
        Baseline   & 60.30 & 64.80   & 51.34  \\
    + $\mathcal{L}_{\mathrm{SFCL}}$    & 63.37 &  67.13   &   54.26 \\
    + $\mathcal{L}_{\mathrm{MHAD}}$  & 64.86 &  69.92  &  55.71 \\
    + $\mathcal{L}_{\mathrm{MHAD}}$ + $\mathcal{L}_{\mathrm{SFCL}}$  & 65.76 &  71.89  &  56.93 \\
    \bottomrule
    \end{tabular}
  \label{tab:ablation_study}
  }
  \vspace{-3mm}
\end{table} 

\begin{figure}[!htbp]
  \centering
    \subfigure[$\alpha=0.3$, $\gamma=1.0$, adjust $\beta$]{
    \includegraphics[width=0.47\linewidth]{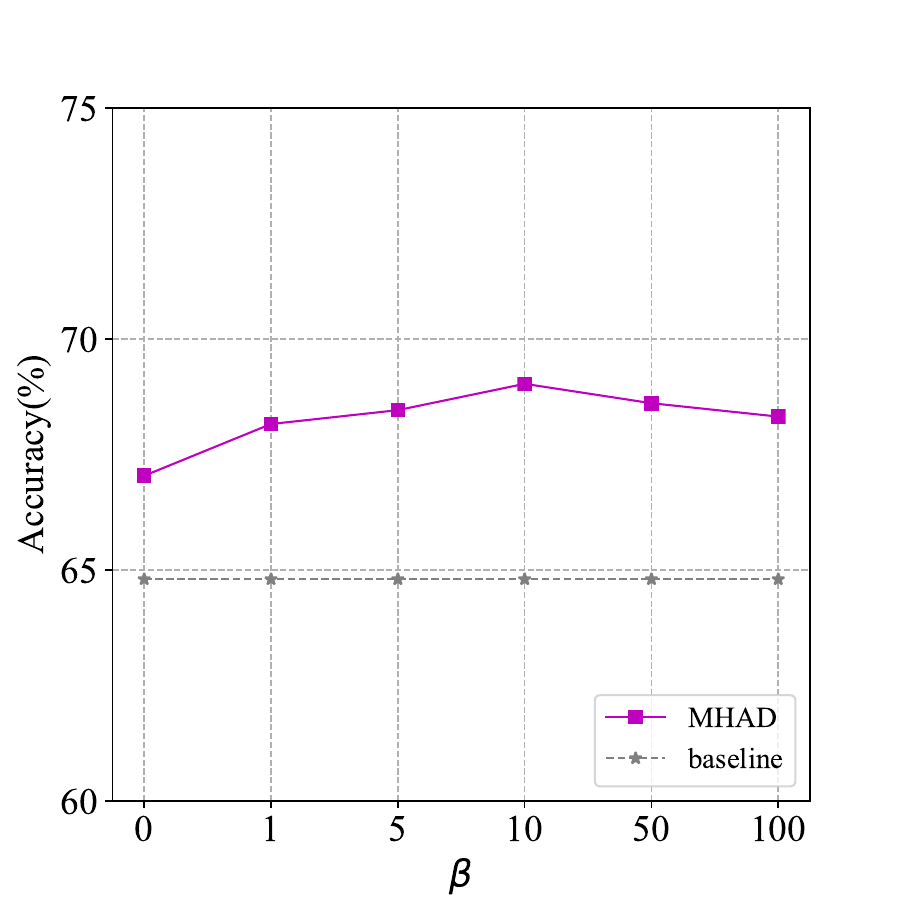}
    \label{fig:ab_a}
  }
    \subfigure[$\alpha=0.3$, $\beta=10$, adjust $\gamma$]{
    \includegraphics[width=0.47\linewidth]{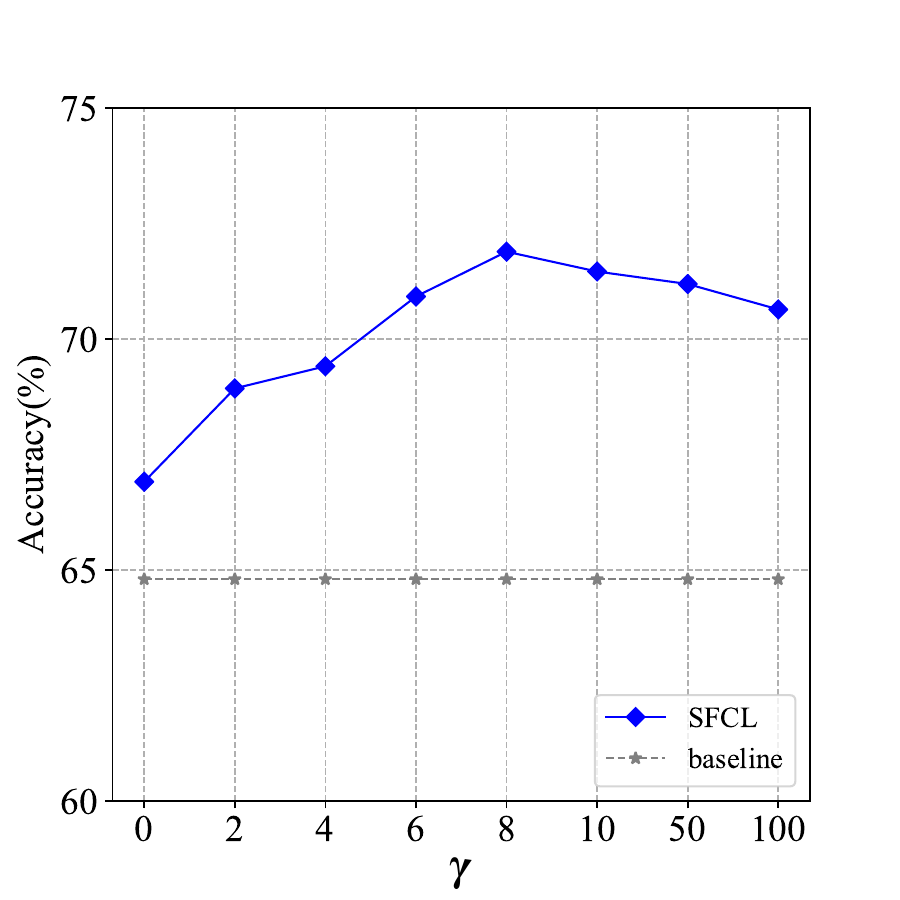}
    \label{fig:ab_b}
  }
  \caption{Effect of hyper-parameter $\beta$ and $\gamma$ on Cars196 dataset.}
  \vspace{-6mm}
  \label{fig:ab}
\end{figure}

\noindent\textbf{Effect of hyper-parameters.}
In our optimization, $\alpha$, $\beta$, and $\gamma$ are the major hyper-parameters for balancing the loss terms in our framework. By adjusting the BN hyperparameter in the interval between 0 to 5, we find that the optimal value of $\alpha$ is 0.3. Then, we investigate the effect of $\beta$ and $\gamma$ on the student ResNet-18 on the Cars196 dataset and show the results in Fig.~\ref{fig:ab}. In Fig.~\ref{fig:ab_a}, we set $\gamma$ as 1.0 and vary $\beta$ from 0 to 100, in which 10 is a reasonable parameter verified by our experiments. Then, we set the optimal value of $\beta$ to 10 and vary $\gamma$ from 0 to 100, in which the student network achieves the best performance when $\gamma$ is set to 8, as shown in Fig.~\ref{fig:ab_b}. 
It is clear that, when using different $\beta$ and $\gamma$, our model stably outperforms the baseline model. The experimental results show that our proposed framework is robust to the different parameters.

\noindent\textbf{Control parameter of attention ganerator.} Due to the parameter $\lambda$ being exploited to control the aggregating of attention and feature maps, we perform a group analysis of this parameter. As shown in Tab.~\ref{tab:param_generator}, we first fix the other parameters, and then the $\lambda$ is set to 0, which indicates that the generator does not exploit the attention mechanism.
And the results on the three datasets achieve 63.88, 69.24, and 54.81, respectively. From the interval 0 to $5e^{-2}$, the effect of the generator rises significantly while the effect of the model decays between $5e^{-2}$ and $9e^{-2}$, in which the reasonable parameter is $5e^{-2}$.
This indicates that the generator needs to be moderate when employing attention. 
When the generator pays too much attention to the attention image, it may destroy the original synthesized images resulting in the degradation of the model.
\begin{table}[htbp]
  \centering
  \vspace{-2mm}
  \caption{The effect of $\lambda$ under different parameters.}
  \vspace{2mm}
  \resizebox{0.7\linewidth}{!}{
    \begin{tabular}{cccc}
    \toprule
    $\lambda$ & Aircraft & Cars196 & CUB200 \\
    \hline
    $0$     & 63.88 & 69.24 & 54.81 \\
    $1e^{-2}$  & 64.32 & 69.87 & 55.44 \\
    $5e^{-2}$   & 65.76 & 71.89 & 56.93 \\
    $7e^{-2}$   & 65.02 & 70.95 & 56.30 \\
    $9e^{-2}$     & 64.23 & 70.36 & 55.81 \\
    \bottomrule
    \end{tabular}%
  }
  \vspace{-4mm}
  \label{tab:param_generator}%
\end{table}%

\noindent\textbf{Order effectiveness of MHA.} We adopt mixed 3-order attention distillation in our method, which is mainly due to the 3-order attention having the ability to pay attention to the context information. It has more information than the 1-order attention. In this section, we conduct experiments to verify the effect of different orders on different FGVC datasets.
As can be seen from Tab~\ref{tab:effect_order}, when exploiting the 1-order attention distillation, we can only achieve 64.31, 69.26, and 56.12 on three datasets. However, when we exploit the 3-order attention distillation, we can improve the scores of 1.5$\%$ on average. 
What exceeded our expectations is the lower effect when 2-order attention was used. 
We believe that the 2-order attention mainly focuses on the global information, including the background information, which confuses the foreground attention and reduces the effect of attention.
\begin{table}[htbp]
  \centering
  \caption{The effect of different orders on different FGVC datasets.}
    \vspace{2mm}
    \resizebox{0.8\linewidth}{!}{
    \begin{tabular}{ccccc}
    \toprule
    Order & Aircraft & Cars196 & CUB200 & Avg.\\
    \midrule
    $R=1$   & 64.31 & 69.26 & 56.12 & 63.23 \\ 
    $R=2$   & 63.12 & 70.35 & 55.06 & 62.84 \\
    $R=3$  & 65.76 & 71.89 & 56.93 & 64.86 \\
    \bottomrule
    \vspace{-4mm}
    \end{tabular}%
    }
    \vspace{-4mm}
  \label{tab:effect_order}%
\end{table}%

\subsection{Architecture of generator} 
As illustrated in Fig.~\ref{fig:framework},  the generator with spatial-wise attention modules is adopted in our experiments. Therefore, we detail the architecture of the generator and attention module as indicated in Tab.~\ref{tab:generator}. Concretely, our generator is isomorphic to DCGAN~\cite{radford2015unsupervised}. However, to facilitate the calculation of the spatial-wise attention module,  we divide the generator into four blocks. At each block, we exploit spectral normalization to normalize the weights of deconvolution, which aims to stabilize the training of the generator. Then, the encoder-decoder spatial-wise attention module is plugged into each block of the generator, in which the indexes of Maxpool are also used in the MaxUnpool to focus on the key position of synthesized features. %

\begin{table}[htbp]
    \centering
    \caption{The \textbf{Left}. Attention Generator Architectures. The noise is mapped to the features which are upsampled to the required image size. The SN denotes the  spectral normalization, while SAM represents spatial-wise attention modules corresponding to the \textbf{Right}.}
    \vspace{2mm}
    \label{tab:generator}
    \setlength\extrarowheight{1pt}
    \resizebox{\linewidth}{!}{
      \begin{tabular}{c|c}
      \Xhline{1pt}
       Attention Generator  &  Spatial-wise Attention Modules \\
      \hline
      FC, Reshape, BN &  1 $\times$ 1 $C  \rightarrow  C/r$ Conv \\
      \hline
      $3\times3$, $ 512  \rightarrow 256  $, Deconv $\uparrow_{2\times}$, & 
      $3\times3$, $C/r  \rightarrow  2C/r$,  Conv, \\
      SN,  LReLU, SAM &  BN, ReLU, Maxpool  \\
      \hline
      $3\times3$, $256  \rightarrow  128$, Deconv $\uparrow_{2\times}$,  & $3\times3$, $2C/r  \rightarrow  4C/r$,  Conv,    \\
      SN, LReLU, SAM &  BN, ReLU \\
      \hline
      $3\times3$, $128  \rightarrow  64$, Deconv $\uparrow_{2\times}$,  & $3\times3$,  $4C/r  \rightarrow  2C/r$, Decov, \\
      SN,  LReLU, SAM  &  BN, ReLU, MaxUnpool \\
      \hline
      $3\times3$, $64  \rightarrow  64$, Deconv $\uparrow_{2\times}$,  &  $3\times3$,  $ 2C/r  \rightarrow  C/r$, Decov, \\
      SN,  LReLU, SAM & BN, ReLU \\
      \hline
      $3\times3$, $64  \rightarrow 3$, Conv, Tanh      & $1\times1$, $C/r  \rightarrow  C$,  Conv, SoftMax \\
      \Xhline{1pt}
      \multicolumn{2}{l}{\small * $C$ is the input channel of each block, while $\mathrm{r}$ is scale scalar.}
      \vspace{-6mm}
      \end{tabular}%
      }

  \end{table}%

\section{Conclusion}
In this paper, we address the data-free distillation for FGVC.
We propose to exploit the generator with spatial attention to synthesize the images with discriminative features.
Then, two effective strategies are exploited to optimize the student by MHAD and SFCL, where MHAD captures the discriminative features with context information and SFCL exploits the high-level semantic features to contrast the variances between the different categories.
Experimental evidence demonstrates that both approaches can improve the performance of the student on FGVC tasks and outperform other data-free distillation approaches to achieve state-of-the-art performance.
\bibliographystyle{ieee_fullname}
\bibliography{main}
\end{document}